\documentclass[num-refs]{wiley-article}

\usepackage{siunitx}

\papertype{Research article}
\paperfield{Journal Advances in Polymer Technology}

\title{Analysis of the fiber laydown quality in spunbond processes with simulation experiments evaluated by blocked neural networks}

\author[1,2]{Simone Gramsch}
\author[1,2]{Alex Sarishvili}
\author[1,2]{Andre Schmeißer}

\affil[1]{Fraunhofer ITWM, Fraunhofer-Platz 1, 67663 Kaiserslautern, Germany}
\affil[2]{Fraunhofer Center for Machine Learning, Germany}

\corraddress{Simone Gramsch, Fraunhofer ITWM, Fraunhofer-Platz 1, 67663 Kaiserslautern, Germany}
\corremail{simone.gramsch@itwm.fraunhofer.de}

\fundinginfo{This work was developed in the Fraunhofer Cluster of Excellence “Cognitive Internet Technologies”.}

\runningauthor{Simone Gramsch et al.}

\usepackage[utf8]{inputenc} 
\usepackage[T1]{fontenc}    
\usepackage{hyperref}       
\usepackage{url}            
\usepackage{booktabs}       
\usepackage{amsfonts}       
\usepackage{nicefrac}       
\usepackage{microtype}      

\usepackage{lineno}

\usepackage{graphicx}

\usepackage{longtable}
\usepackage{colortbl}

\usepackage{amsmath, amssymb}
\usepackage{textcomp}
\usepackage{siunitx}
\usepackage{enumitem}

\begin{document}

\begin{frontmatter}
\maketitle

\begin{abstract}
We present a simulation framework for spunbond processes and use a design of experiments to investigate the cause-and-effect-relations of process and material parameters on the fiber laydown on a conveyor belt. The analyzed parameters encompass the inlet air speed and suction pressure, as well as the E modulus, density and line density (titer) of the filaments. The fiber laydown produced by the virtual experiments is statistically quantified and the results are analyzed by a blocked neural network. This forms the basis for the prediction of the fiber laydown characteristics and enables a quick ranking of the significance of the influencing effects. We conclude our research by an analysis of the nonlinear cause-and-effect relations. Compared to the material parameters, suction pressure and inlet air speed have a negligible effect on the fiber mass distribution in (cross) machine direction. Changes in the line density of the filament have a 10 times stronger effect than changes in E modulus or density. The effect of the E modulus on the throwing range in machine direction is of particular note, as it reverses from increasing to decreasing in the examined parameter regime.

\end{abstract}
\end{frontmatter}

\section{Introduction}
Over the last decade, the annual growth rate for global nonwovens production has averaged 5.7 percent \cite{edana:url}. The reason for this is the wide range of applications for nonwovens: They range from hygiene and medicine to construction, home furnishings, clothing, and automobiles (see, e.g., \cite{edana:urlmarket}). However, the versatility of the fields of applications has the disadvantage that processes should be adapted to the subsequent applications of the nonwovens as far as possible.

Due to their big impact on the nonwovens market -- \cite{reportsanddata:2019} estimates the valuation for the spunbond nonwovens market at USD 11.50 billion in 2018 -- spunbond processes experience constant attention in textile research. The review article \cite{lim:2010} gives not only a detailed summary of the technical principles and the current market trends of spunbond processes, but also covers the history of the spunbond technology. Apart from that, current publications focus more on applications of spunbonded fabrics for composites or filtration. E.g. in \cite{kiselev:2013} results of experimental studies are presented, where the porosity of the spunbonded fabrics in composites is analyzed. The authors of \cite{gheryani:2017} deal also with spunbonded nonwovens in composites, but focuses on their application as interleaves and the enhanced resulting toughness of the composite.  
In the application of spubonded nonwovens for filter media, the bicomponent nonwovens in particular play an important role. Filtration properties for bicomponent spunbonded nonwovens are evaluated in \cite{heng:2015}. In \cite{maltha:2012}, spunbonded bicomponent nonwovens are also used in order to achieve better filter medium efficiency of cabin air filters, while \cite{yeom:2011} focuses on aerosol filtration properties of PA6/PE islands-in-the-sea bicomponent spunbond fabrics. 

In addition to the research work that deals with applications of nonwovens, there is ongoing general experimental research of spunbond systems. Hereby, the most important question is how the process conditions influence the nonwoven web quality with respect to mechanical properties like tensile strength or stiffness as well as other properties like crystallinity. For example in \cite{bhat:2004} the influence of thermal bonding conditions on the structure of spunbonded nonwovens is analyzed. In \cite{fedorova:2007}, the feasibility of using of islands‐in‐the‐sea fibers in the spunbond process to produce relatively high strength micro‐ and nanofiber webs is explored. The hydraulic properties of electrospun fiber webs and spunbond nonwoven fabrics were compared in \cite{hong:2006}. We refer to \cite{nanjundappa:2005} where a broad range of process variables was studied to investigate the relation between process and properties. Current discussions about sustainability lead to an increasing interest in spunbond processes with polymers from renewable resources like PLA, see, e.g., \cite{shim:2016}.

According to the current state of art in industry, process and product design is carried out by trial and error on production lines, which is time-consuming and cost-intensive. Scaling up experiment-based optimization results from pilot lines to production plants can be misleading due to the nonlinearity of the influencing factors. In particular the air stream is highly sensitive to varying process conditions and, hence, the turbulent effects of the air within the spunbond processes. In order to support the experimental approach of designing spunbond processes with respect to customer-specific needs, theoretical analyses and simulation methods become a key technology. For example in \cite{kanai:2018} a physical model is developed describing the fiber properties like fiber diameter, fiber speed, strain rate, stress, temperature and crystallinity dependent on the process conditions. Mathematical models that also describe the laydown of the fibers on the belt can be found in \cite{klar:2009} or \cite{wegener:2015}. A description of the software that implements these models is found here \cite{gramsch:2015}.

In this paper, we start by presenting a simulation framework for spunbond processes. With a design of experiments we study the influencing effects of process and material parameters of the fiber laydown on a conveyor belt. We train a feed-forward neural network in order to study a quality criterion of the fiber laydown structure. We conclude with an analysis of the cause-and-effect relations and a ranking of the input effects on the fiber laydown. 

\section{Principles of nonwoven production processes and simulation framework of spunbond processes} \label{sec:02}
\begin{figure}[!htb] \centering
\includegraphics[width=0.9\textwidth]{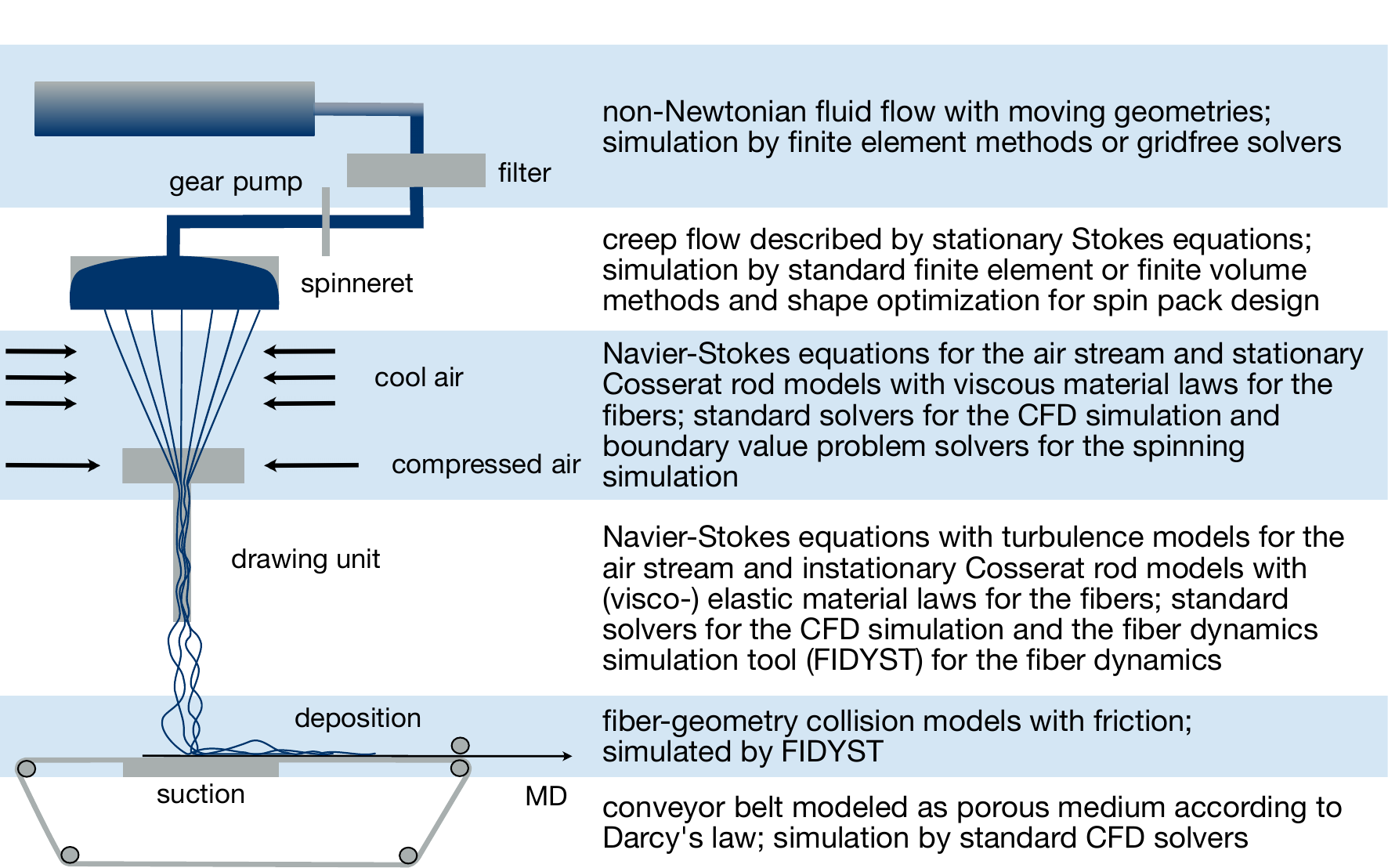}
\caption{Sketch of a spunbond process with physical models for single process steps; the enumeration of the mentioned simulation methods is not complete, but should be understood as examples.} \label{fig:01}
\end{figure}

Spunbond processes follow in principle the process steps extrusion, spinning, drawing, laying down, and bonding. Figure~\ref{fig:01} shows a sketch of a typical spunbond process used for the production of nonwovens. On the right hand side of Fig.~\ref{fig:01} corresponding physical models describing single process steps are quoted and the currently dominating simulation techniques are listed. Hereby, the enumeration of simulation methods is not complete, but should only be understood as examples.

More precisely, a melted polymer (e.g. polypropylene or polyester) is extruded, filtered, and transported to a so-called spin pack. In the spin pack the melt is forced through hundreds of holes called nozzles or spinnerets. Streams of viscous melt exit the spinnerets and form viscous fibers. The fibers are cooled and stretched by an air flow coming from the side. Then they are driven by compressed air through a channel, the so-called drawing system. After they have left the drawing channel, turbulent air streams entangle them. Finally, they lay down on the conveyor belt and form a random web. Suction underneath the belt prevents the fibers from rebounding. The random web is transported away for further post-processing steps like mechanical or thermal bonding. We refer to \cite{albrecht:2002} for further details.

\begin{figure}[!htb] 
\begin{tabular}{lr}
\includegraphics[width=0.45\textwidth]{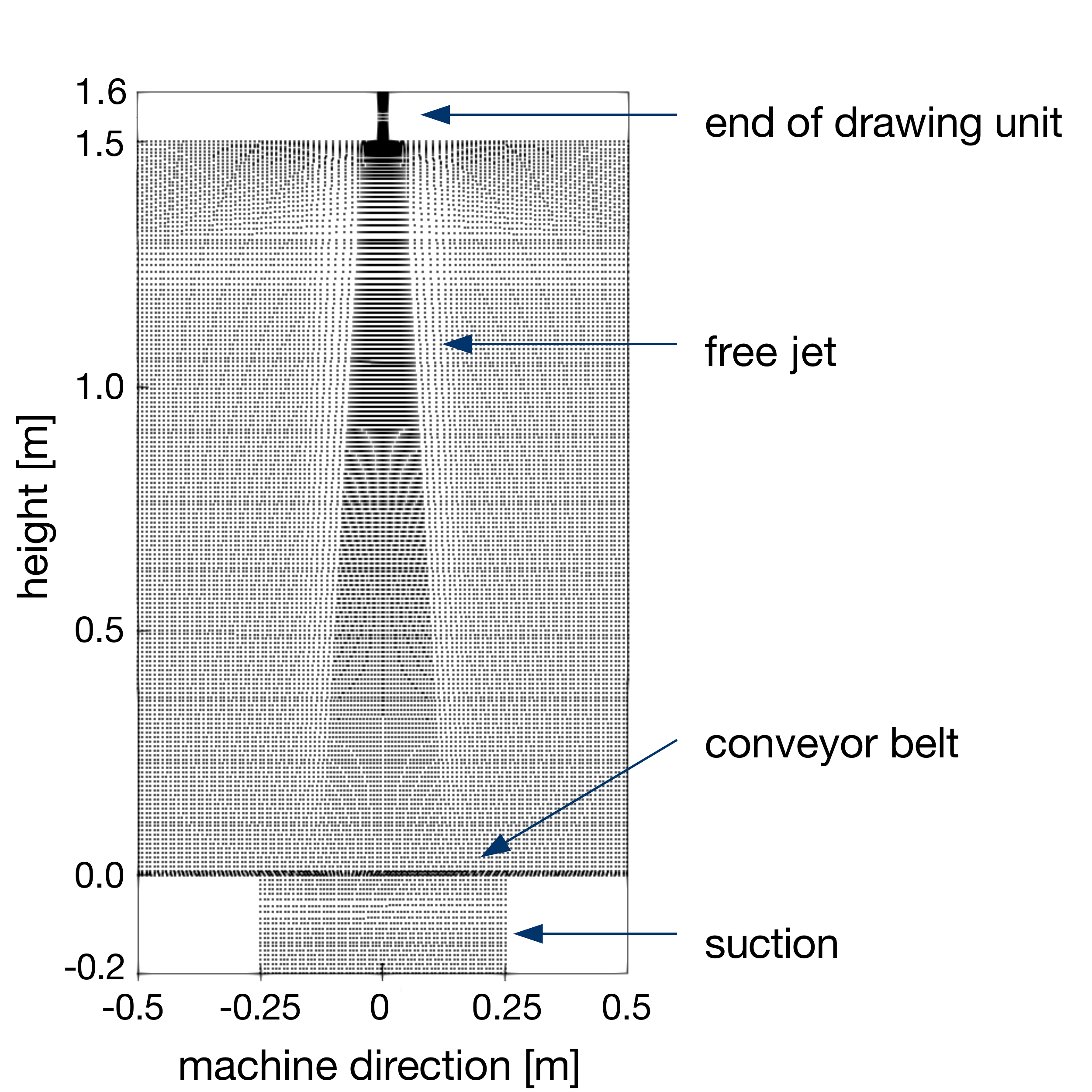} &
\includegraphics[width=0.52\textwidth]{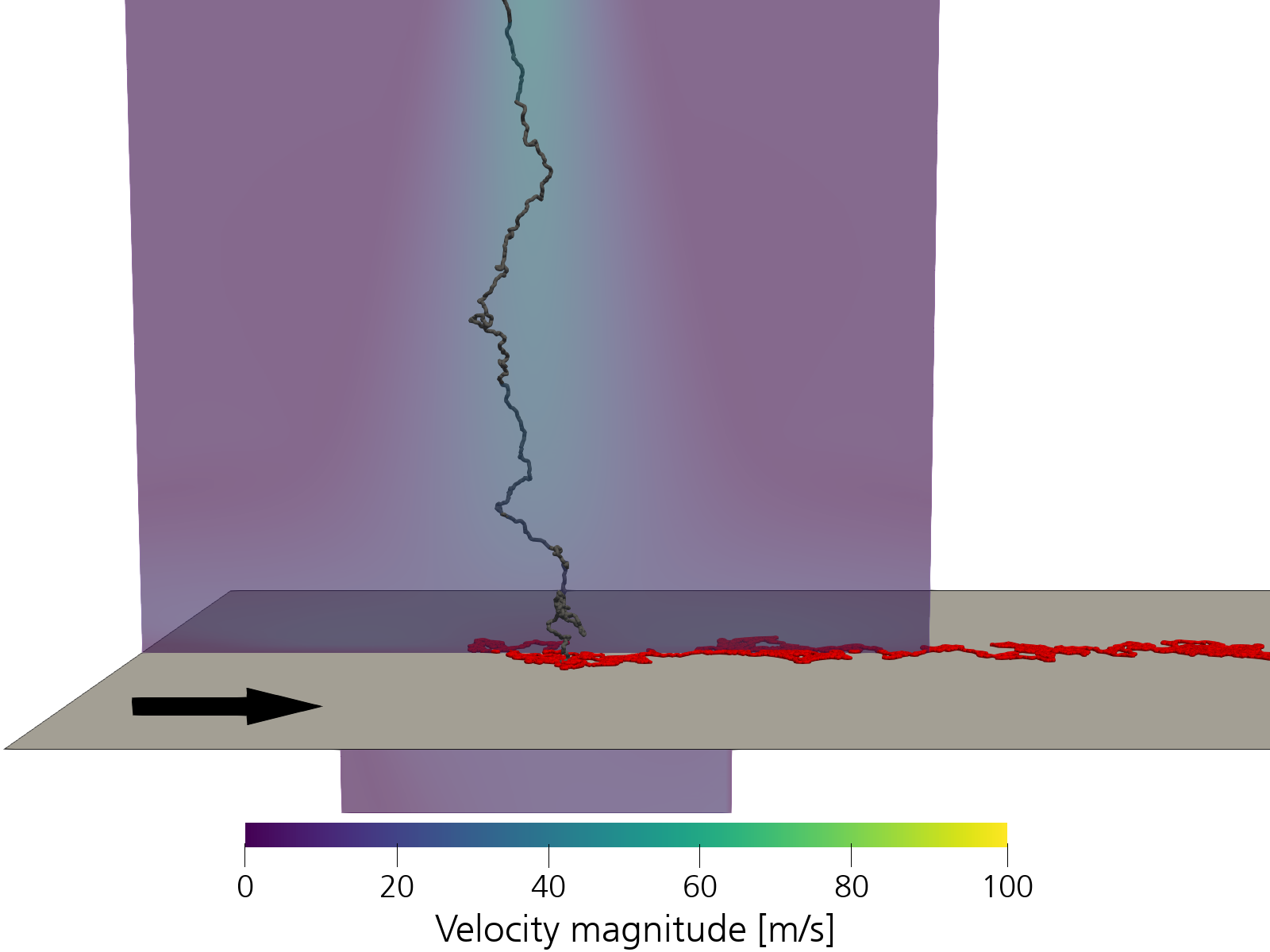}
\end{tabular}
\caption{Left: Mesh of the used simulation domain for the air stream; the domain covers mainly the free jet in spunbond processes from the end of the drawing unit to the conveyor belt with suction underneath. Right: exemplary result of a fiber dynamics simulation; the fiber parts in the air are colored black, the deposited fiber parts on the conveyor belt red. The air stream is visualized as a slice.}
\label{fig:mesh+fidyst}
\end{figure}

In this study we put our focus on the entangling and laydown phase of fibers within spunbond processes. Hereto, we simulate the air stream starting at the end of the drawing unit to the suction beneath the conveyor belt. Fig.~\ref{fig:mesh+fidyst} shows the used simulation domain with geometry dimensions. As physical model for the air stream we use the \emph{Reynolds-averaged Navier-Stokes equations} (or RANS equations) with a so-called k-$\varepsilon$-turbulence model. The moving conveyor belt is modelled as a porous medium according to Darcy's law. All simulations of the air stream (CFD simulations) are performed as stationary 2d simulations assuming periodic boundary conditions in cross machine direction (CD). This corresponds to a single-row, multi-column spinneret. The following considerations therefore hold true only for the center regions of a spunbond process, i.e. the inner columns, and can not be used to investigate the boundary effects of the fiber laydown. Of course, the study can be extended in the future in order to analyze the boundary effects, as well as to analyze multi-row processes.

For the fiber dynamics driven by the turbulent air we use a modeling framework based on the theory of \emph{Cosserat rods}, where the fiber is modeled as a one-dimensional object, as the fiber diameter is negligibly small compared to its length. The fiber is modeled by a curve $\mathbf{r}$  describing its centerline and an orientation of its cross-section, given by a set of directors $\mathbf{d}_i$ forming an orthogonal basis. This general framework consists of equations for the fiber's kinematics and dynamics, and is complemented with a material model and geometry model specific to the spunbond process, see \cite{wegener:2015}. The material model assumes elastic, inextensible behavior of the fiber in the lower part of the production process, whereas the geometry model assumes a circular cross-section of the fiber with a constant radius. Using these assumptions we arrive at a simplified string model for the fiber from the more general Cosserat model by lengthy derivations (cf. \cite{wegener:2015}), given as:
\begin{linenomath*} \begin{align}
\begin{split}
(\rho A) \partial_{tt} \mathbf{r} & = \partial_s (T \partial_s \mathbf{r} - \partial_s((EI)\partial_{ss} \mathbf{r})) + \mathbf{f}_\text{ext}, \\
\|\partial_s \mathbf{r}\| & = 1. \label{eq:fideq}
\end{split}
\end{align} \end{linenomath*}
Here, the fiber centerline $\mathbf{r} : (s_a, s_b) \times \mathbb{R}^{+} \to \mathbb{R}^3$ is a function of the material parameter $s$ as well as time $t$, $(\rho A)$ is the line density [\si{kg/m}] (also called titer) of the fiber, $(EI)$ is the bending stiffness [\si{Nm^2}], $T$ is the tangential contact force [\si{N}], and $\mathbf{f}_\text{ext}$ includes all external line forces [\si{N/m}]. Note that the material parameters $(\rho A)$ and $(EI)$ are prescribed along the fiber, i.e., as input of the simulation, whereas the centerline $\mathbf{r}$ and contact force $T$ are computed by the simulation, i.e. output.

The line force $\mathbf{f}_\text{ext}$ is the sum of all external forces, including aerodynamic forces, gravity, and contact forces arising from the deposition of the fibers onto the conveyor belt. Modeling these forces is crucial for the correct simulation of the fibers, see \cite{marheineke:2011} for the aerodynamic forces, \cite{gramsch:2014} for the contact forces and numerical regularization as in \cite{schmeisser:2015}. With this formulation and modeling the conveyor belt as a planar object, the contact forces take a simple form as
\begin{linenomath*} \begin{align}
\mathbf{f}_\text{contact} =& \lambda \mathbf{n}_b, \label{eq:contact_force}\\
(\lambda = 0 \wedge \mathbf{n}_b (\mathbf{r} - \mathbf{x}_b) \ge 0) \qquad \vee& \qquad (\lambda > 0 \wedge \mathbf{n}_b (\mathbf{r} - \mathbf{x}_b) = 0), \label{eq:contact_constraint}
\end{align} \end{linenomath*}

where the belt is given as a plane with normal $\mathbf{n}_b$ through the point $\mathbf{x}_b$, and the magnitude $\lambda$ of the force is computed as a Lagrange multiplier to the non-penetration constraint \eqref{eq:contact_constraint}. Further, this is completed by a friction model.

For the simulation the resulting partial differential equations are discretized in time $t$ and space $s$ and then integrated using the implicit Euler method. This requires solving a non-linear system of equations for each time step, which in turn is solved using Newton's method. More details about the discretization scheme and an industrial application of this simulation method for the dynamics of staple fibers can be found in \cite{gramsch:2016}.

\section{Design of experiments and corresponding CFD and fiber laydown simulation results} \label{sec:03}

The model of the fiber dynamics and laydown requires a description of the external forces acting on the fibers as well as specification of their material properties. Hereto, we use the geometry with the mesh specified in Fig.~\ref{fig:mesh+fidyst} as starting point for all following trials. We perform a series of simulations using a design of experiments (DoE) in two steps: first we create a base data set of the effects of the material parameters, which is then augmented by the CFD parameters. Thus, for the first part of the DoE, we vary the material properties while using a fixed CFD data set corresponding to the central values $v = \SI{100}{m/s}$, $p = \SI{100}{Pa}$, and fixed process parameters. Hereby, $v$ denotes the inlet speed of the air at the end of the drawing unit, while $p$ denotes the pressure of the suction under the conveyor belt (compare Fig.~\ref{fig:mesh+fidyst}). The numerical simulation parameters are also kept constant, i.e., we use a discretization time $\Delta t = \SI{1}{\micro s}$, discretization length $\Delta s = \SI{1}{mm}$, and total simulation time $T = \SI{2}{ s}$, producing \SI{160}{m} of fiber length at a spinning speed of $u = \SI{80}{m/s}$. For the material parameters, we use a Latin Hypercube scheme with 100 points in the following ranges:
\begin{enumerate}
\item E modulus in the range \SIrange{10}{30}{GPa},
\item density in the range \SIrange{900}{1200}{g/cm^3}, and
\item line density (titer) in the range \SIrange{2.83}{4.53}{dtex}.
\end{enumerate}

\begin{figure}[!htb] \centering
\includegraphics[width=0.49\textwidth]{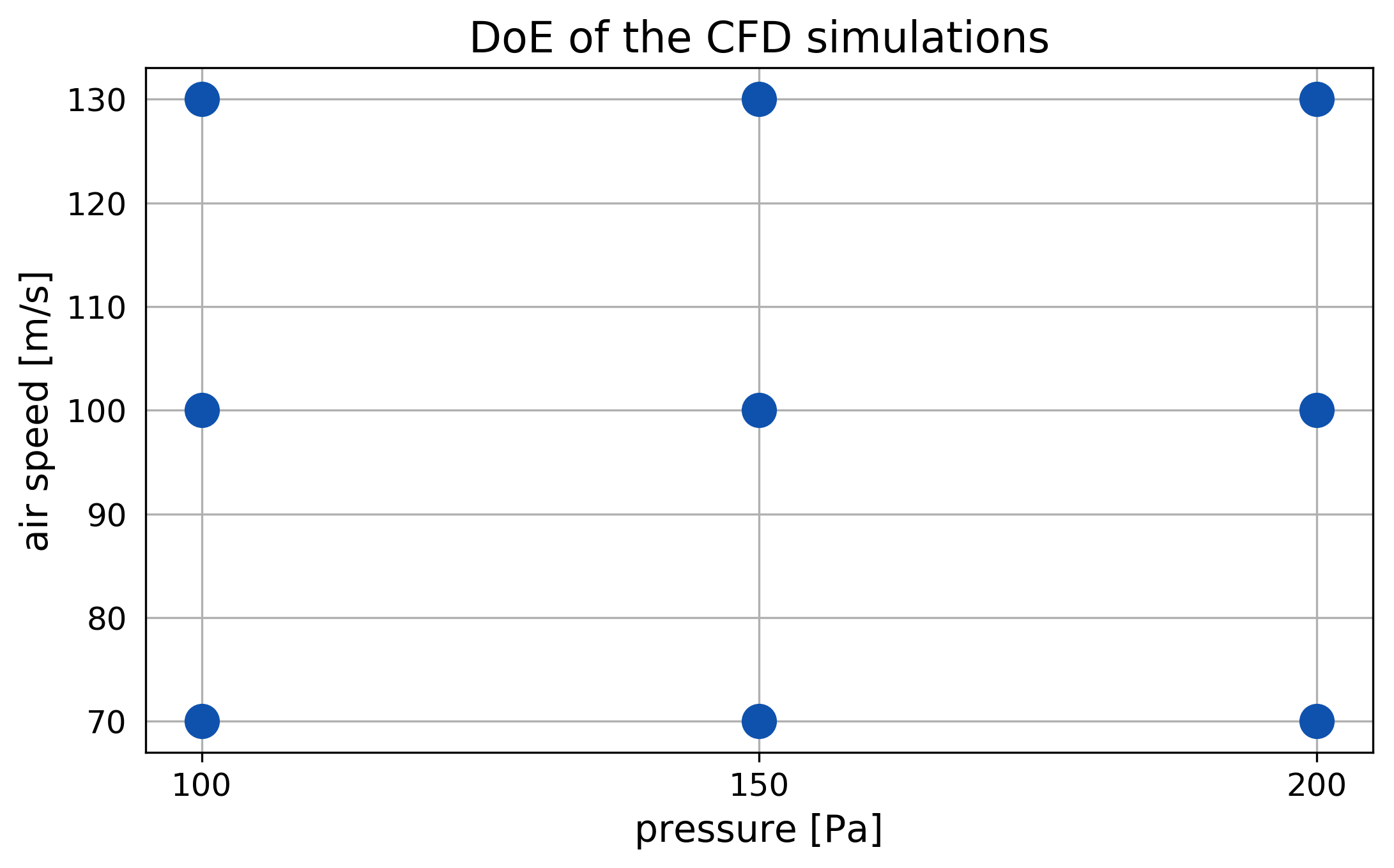}
\includegraphics[width=0.49\textwidth]{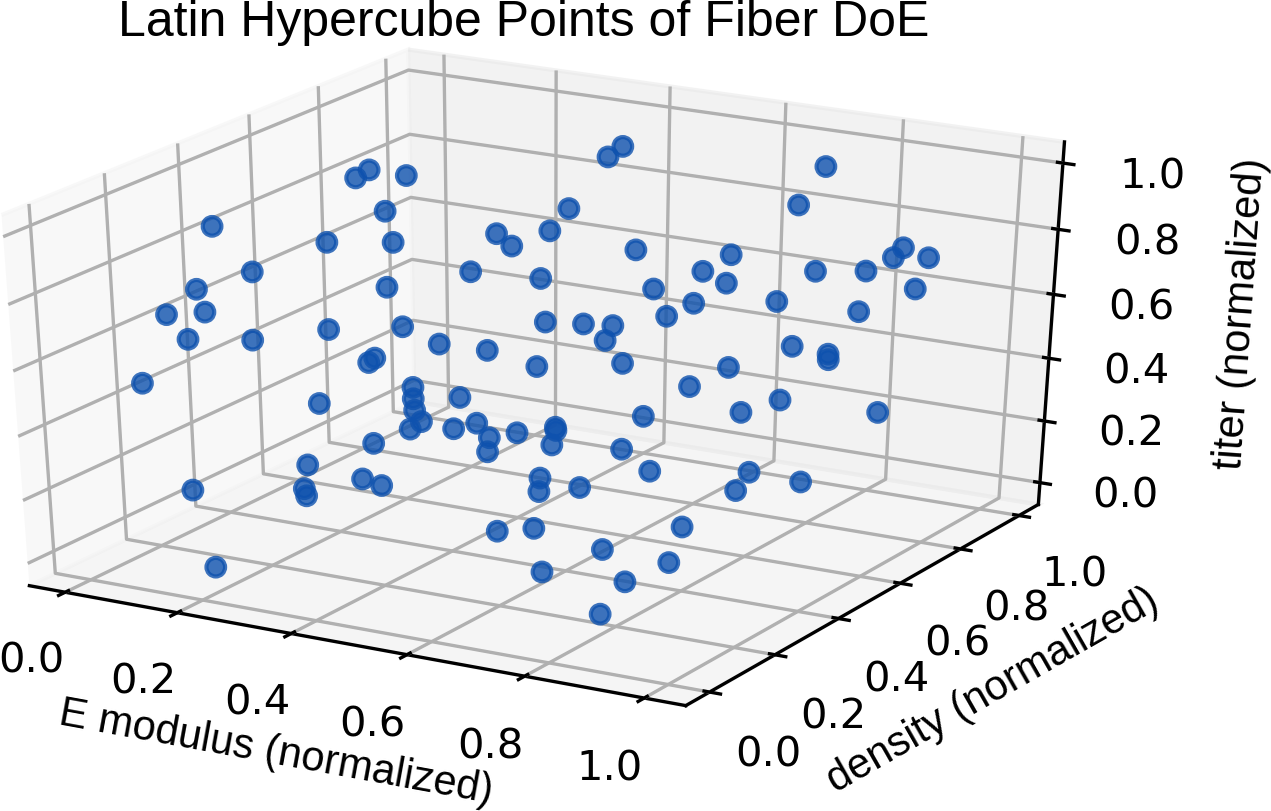}
\caption{Left figure: design of experiments for the CFD simulations. Right figure: distribution of the 100 Latin Hypercube points of the fiber DoE: the original range of the E modulus is \SIrange{10}{30}{GPa}, the density range \SIrange{900}{1200}{g/cm^3}, and the line density (titer) range \SIrange{2.83}{4.53}{dtex}. A simulation has been performed for each parameter combination corresponding to the points in the cube. For better visualization the values are plotted in a normalized range between \num{0} and \num{1}. } \label{fig:doe1}
\end{figure}

The distribution of these material parameters is shown in Fig. \ref{fig:doe1}, where we have used a normalized scale for better visualization.

\begin{figure}[!htb] \centering
\includegraphics[width=0.3\textwidth]{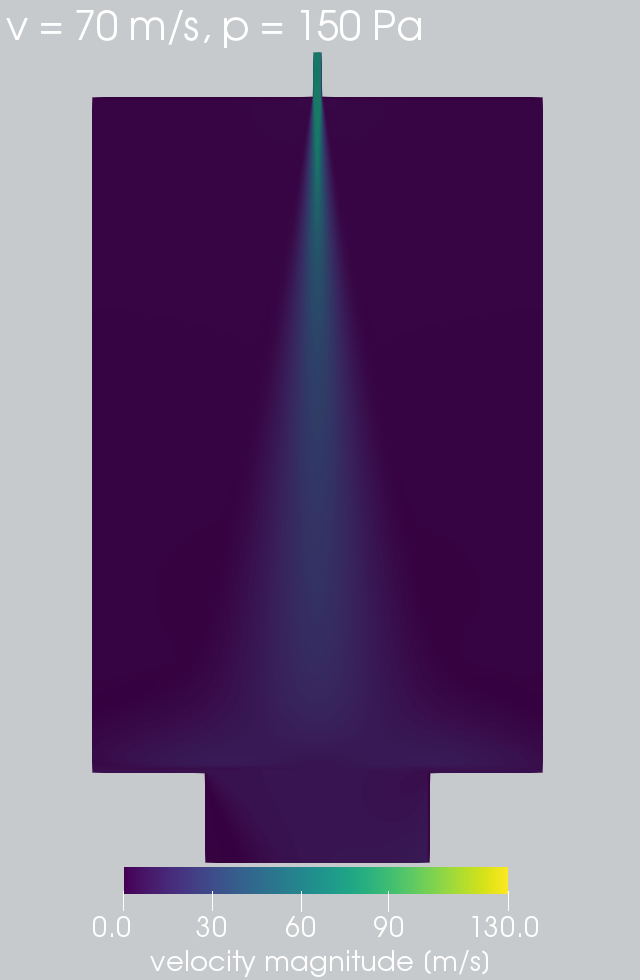} 
\includegraphics[width=0.3\textwidth]{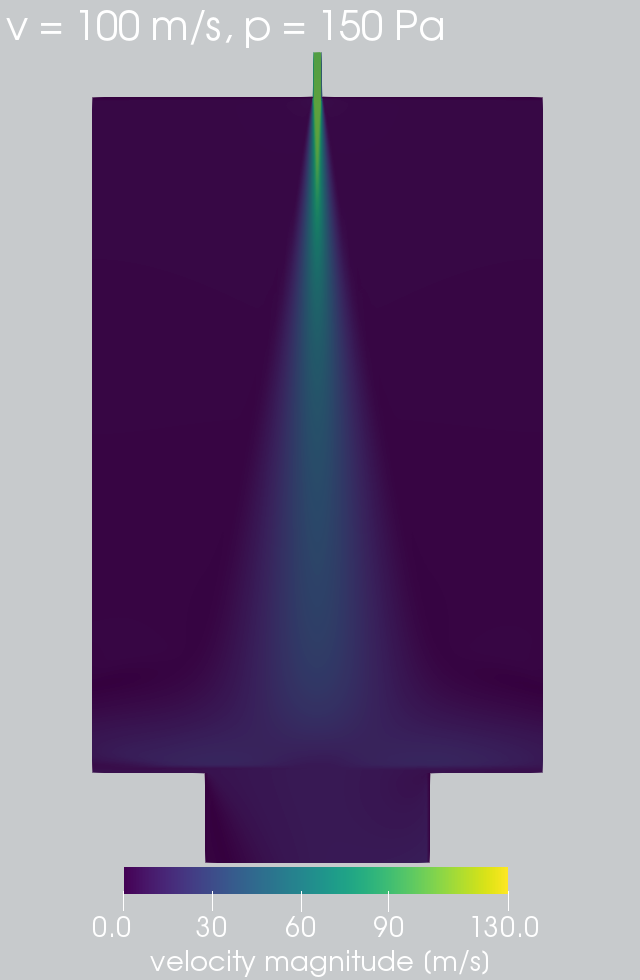}
\includegraphics[width=0.3\textwidth]{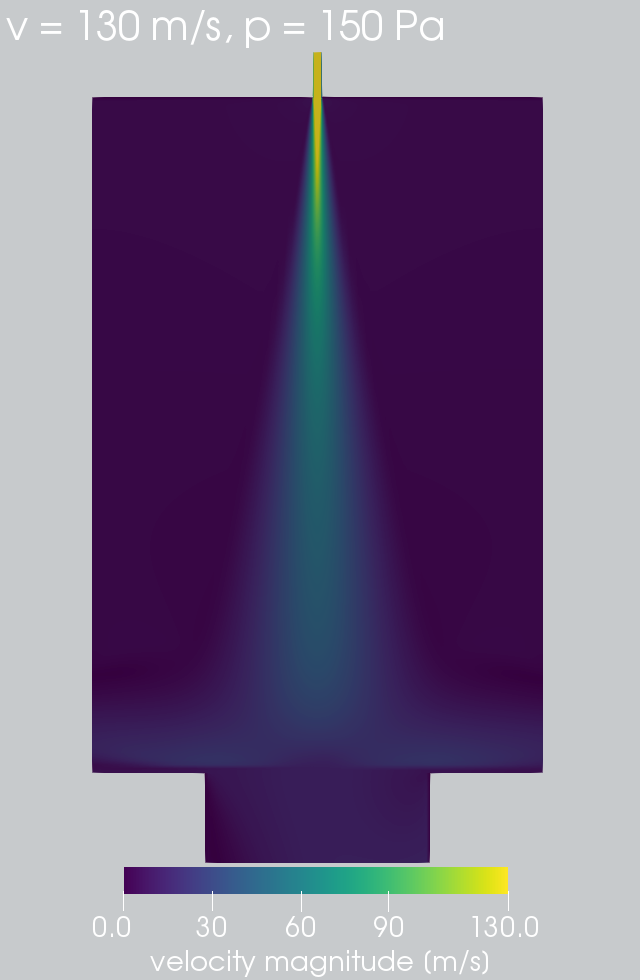}
\caption{Comparison of air speeds for varying inlet speeds at the end of the drawing unit. From left to right: \SIlist{70; 100; 130}{m \per  s}, while the absolute pressure of the suction is fixed at $p = \SI{150}{Pa}$.}
\label{fig:velMag}
\end{figure}

\begin{figure}[!htb] \centering
\includegraphics[width=0.3\textwidth]{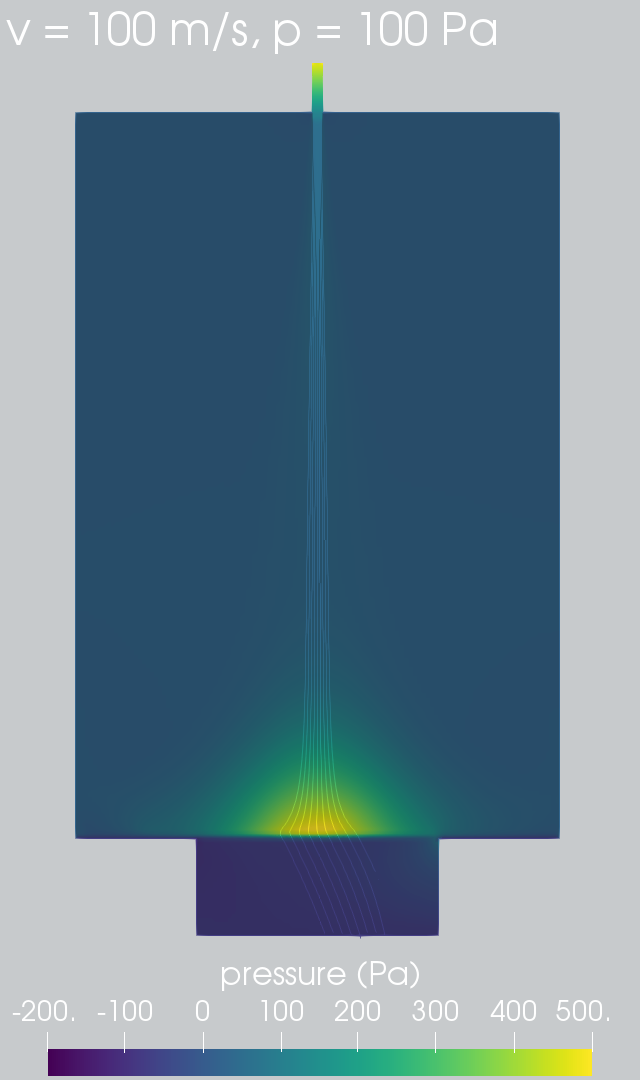} 
\includegraphics[width=0.3\textwidth]{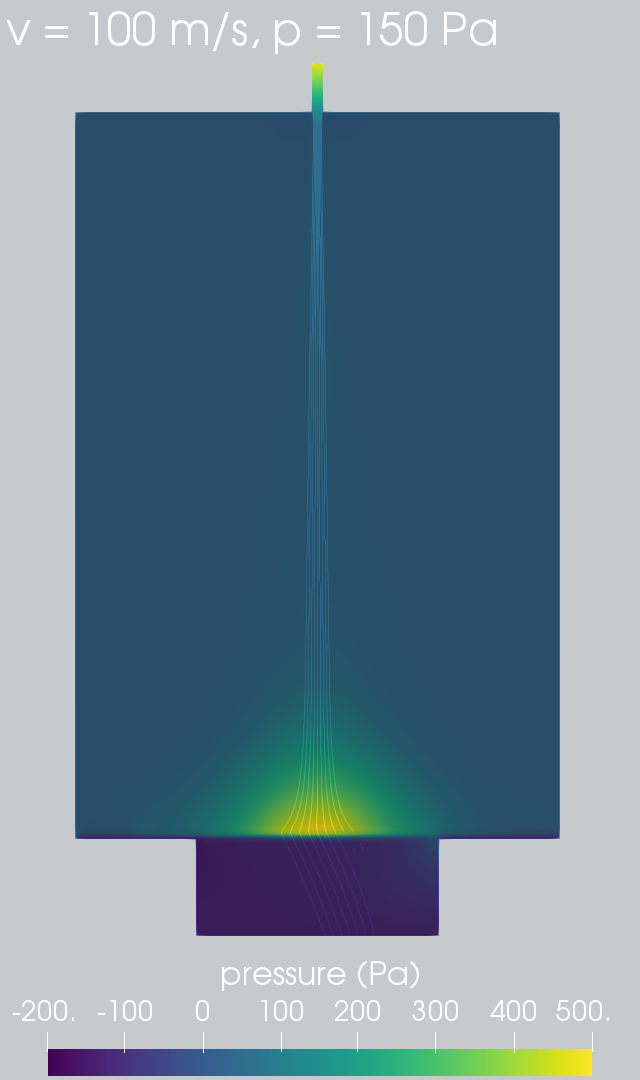}
\includegraphics[width=0.3\textwidth]{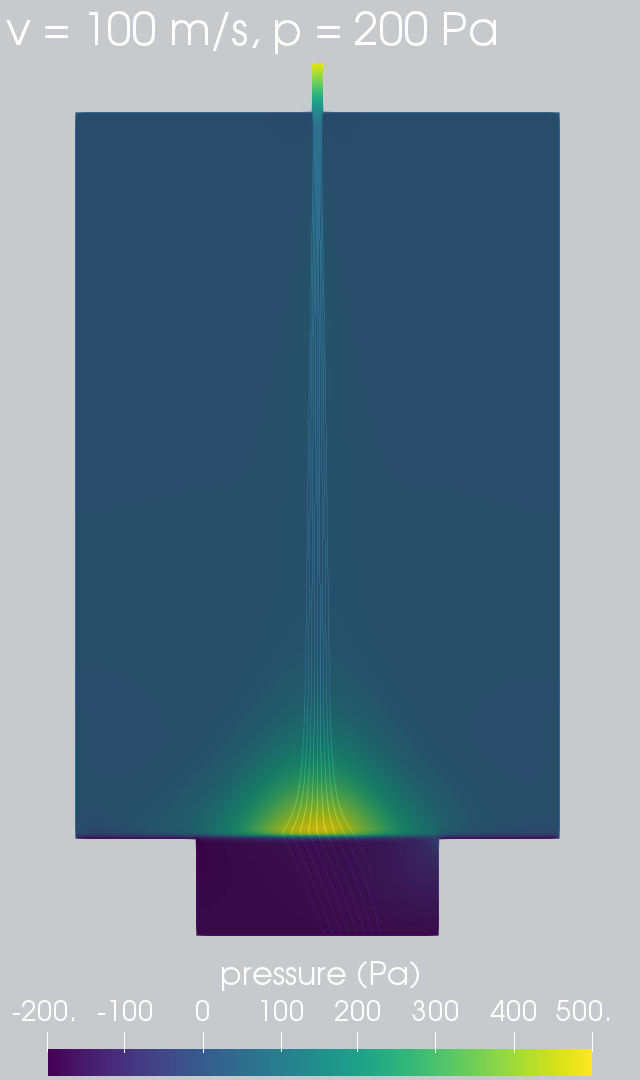}
\caption{Comparison of the pressure in the entangling zone for varying suction pressure. From left to right: \SIlist{100; 150; 200}{Pa}, while the inlet speed at the end of the drawing unit is fixed at $v = \SI{100}{m \per s}$.}
\label{fig:pressure}
\end{figure}

In a second step, we augment the DoE by an additional \num{105} simulations where all five parameters are varied, i.e., the two process parameters of the CFD simulation and the three material parameters of the fiber dynamics simulation. Because of to the very high computation costs of the CFD simulation, again only the nine discrete values as given in Fig.~\ref{fig:doe1} are used. Due to the nonlinearity of the airflow with regard to the input parameters, a higher number of discrete samples could further improve the accuracy of the results, but at significant costs. In Fig.~\ref{fig:velMag} the simulation results of the air speed are presented for varying input conditions of the air speed at the end of the drawing channel, while Fig.~\ref{fig:pressure} shows simulations results of the pressure. Also, the three material parameters are no longer continuously sampled but instead correspond to the normalized values \numlist{0.1; 0.5; 0.9} each. Hereby, we choose \num{0.1} instead of \num{0} and \num{0.9} instead of \num{1} in order to avoid the extreme edges of the Latin Hypercube design. With this augmentation we get a data set of \num{205} simulation settings which gives a good sampling of the interior of the 5-dimensional input parameter range.

\begin{figure}[!htb] \centering
\includegraphics[width=\textwidth]{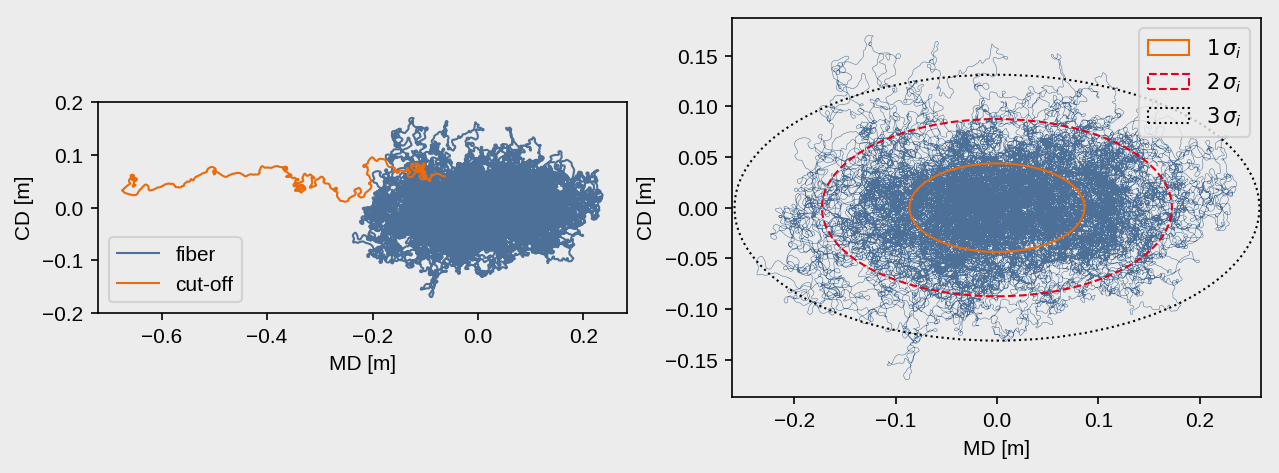}
\caption{(left) Backtracked fiber laydown with tail (orange) caused by process initialization, (right) standard deviations of throwing range for laydown with tail cut off.} \label{fig:laydown}
\end{figure}

From the fiber dynamics simulation, we want to judge the quality of the resulting nonwoven. As the simulation produces endless filaments, the part of the fiber that is still in air is discarded and only the laydown is considered, i.e., the part of the fiber that has already been deposited onto the belt, see Fig. \ref{fig:laydown}. Of this laydown a "backtracked" version is computed where the transport of the fiber along the belt is subtracted. Thus, we reconstruct a distribution of the fiber below the spinneret. Since the simulation is initialized with a short fiber being spun into a long filament, the fiber initially can have a different behavior while the free end is in air compared to the real process where parts of the fiber are already laid down. We also discard the first part of the fiber laydown to remove this outlier effect, see Fig. \ref{fig:laydown}.

We extract three statistical parameters $\sigma_1$, $\sigma_2$ and $A$ from this laydown which represent the overall stochastic structure of the laydown. Assuming the fiber laydown corresponds to a two-dimensional normal distribution, we compute the standard deviations $\sigma_1$ and $\sigma_2$ of the throwing ranges in MD and CD direction. Additionally, we compute a parameter $A$ that corresponds to the stochasticity of the fiber deposition, where small values of $A \to 0$ correspond to a deterministic deposition and $A \to \infty$ to a completely stochastic process. The reader is referred to \cite{klar:2009} for details of the computation of $A$.

Using these three statistical properties of the simulated representative fibers, we can parametrize a stochastic surrogate model based on a Wiener process, which in a further step allows us to compute a full virtual nonwoven sample consisting of thousands of fibers. This sample can then be analyzed with regard to homogeneity, base weight distribution, etc., giving a measure of quality. However, in the following, we directly use the parameters extracted from the fiber simulation as a proxy for predicting quality, i.e., we consider these as the output values of our simulations which we want to predict given a set of input values.

\section{Study of the influencing parameters by blocked neural networks} \label{sec:04}
Overall goal of this study is to analyze the influence of the process parameters spinning speed and pressure of the suction as well as of the material parameters E modulus, density, and line density on the fiber laydown characterized by $\sigma_1$, $\sigma_2$, and $A$. For simplicity, we denote the process/material parameters as input variables of the spunbond system, while the fiber laydown characteristics are denoted as output variables. A first, rough look at the resulting simulation data shows that the effect of the input variables is of nonlinear nature. Hence, we have to use a generalized regression model for the analysis.  

A good choice for generalized regressions models are feed-forward neural networks. The main advantage of neural networks to other regression models is their universal approximation framework realized by their special architecture. Since many neural networks are constructed by single neurons, we briefly give a short introduction of the mathematical concept of neurons or so-called perceptrons.  

\begin{figure}[!htb]\centering
\includegraphics[scale=0.565]{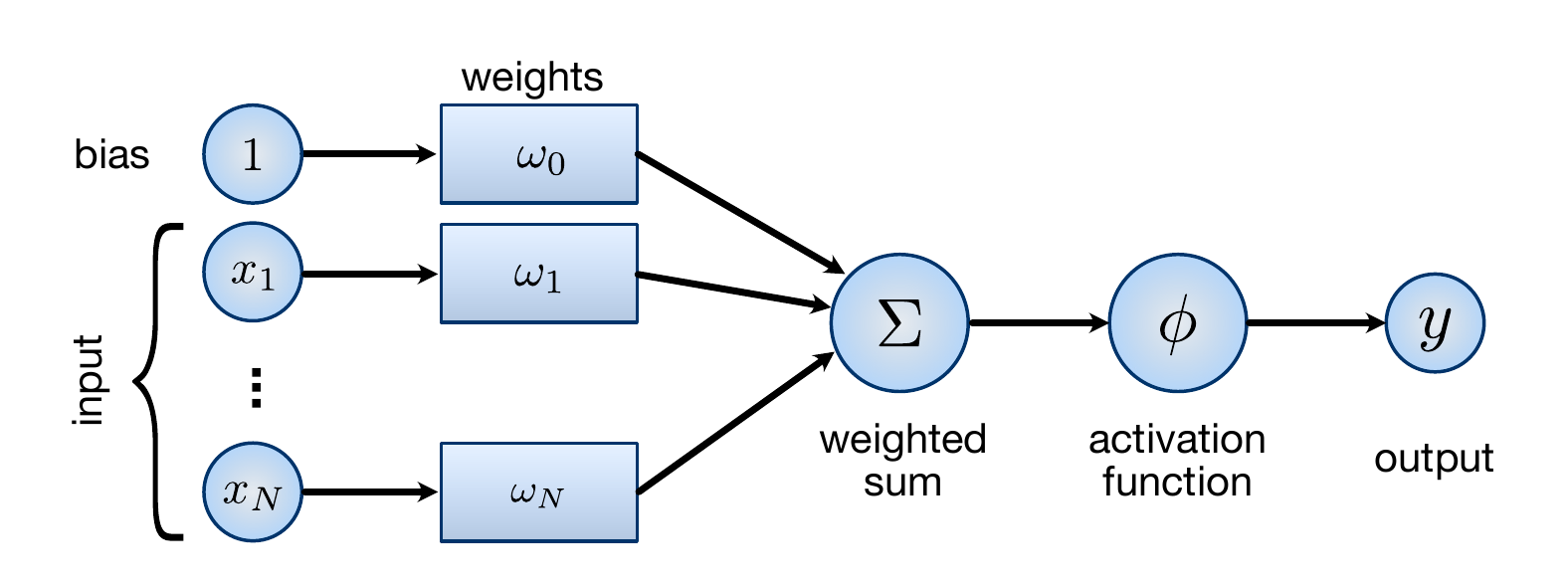}
\includegraphics[scale=0.565]{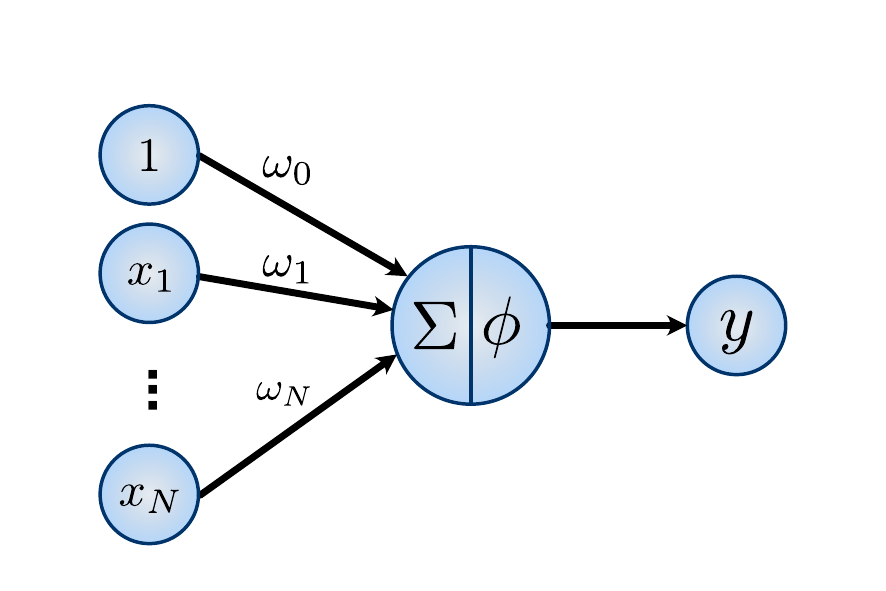}
\caption{Graphical representation of a single neuron/perceptron used to construct neural networks. The neuron consists of $N+1$ weights together with a so-called activation function that is applied to the weighted sum of the inputs. The left figure shows the perceptron in a detailed diagram, while the right representation is usually used if the perceptron is part of a neural network. Hereby, the weighted sum is combined with the application of the activation function in one circle.}
\label{fig:perceptron}
\end{figure}

A single neuron consists of $N$ inputs $x_{i}\in\mathbb{R}$, $i=1, \ldots, N,$ with corresponding weights $\omega_{i}, i=1,\ldots,N$. Sometimes a bias $\omega_{0}$ corresponding to a permanent input of \SI{1} is added. A so-called activation function is attached to the definition of a neuron. Since we consider single neurons, we regard only one output variable $y\in\mathbb{R}$. Together these parts form the neuron by implementing the following two rules: First, the weighted sum of all input variables is computed, i.e., $a = \sum_{i=0}^{N} \omega_i x_i$ (including the bias), then the activation function $\phi$ is applied, i.e., 
\begin{linenomath*} \begin{equation*}
y = \phi(a) = \phi\left(\omega_0 + \sum_{i=1}^{N} \omega_i x_i\right).
\end{equation*} \end{linenomath*}
Graphically, neurons can be presented by one of the following two typical network diagrams as depicted in Fig.~\ref{fig:perceptron}.

In \cite{funa:1989} or \cite{hornik:1989} it is shown that fully connected neural networks are able to approximate arbitrary continuous functions with arbitrary accuracy. Furthermore, in \cite{hornik:1990} it is proven that neural networks with appropriate smooth activation functions are able to approximate the derivatives of the regression functions -- useful, e.g., for optimization. Since our main focus of this paper is to gain more insight into the cause-and-effect-relations from process/material parameters to the fiber laydown characteristics, we use a \emph{blocked neural network} approach. 

\begin{figure}[!htb] \centering
\includegraphics[scale=0.65]{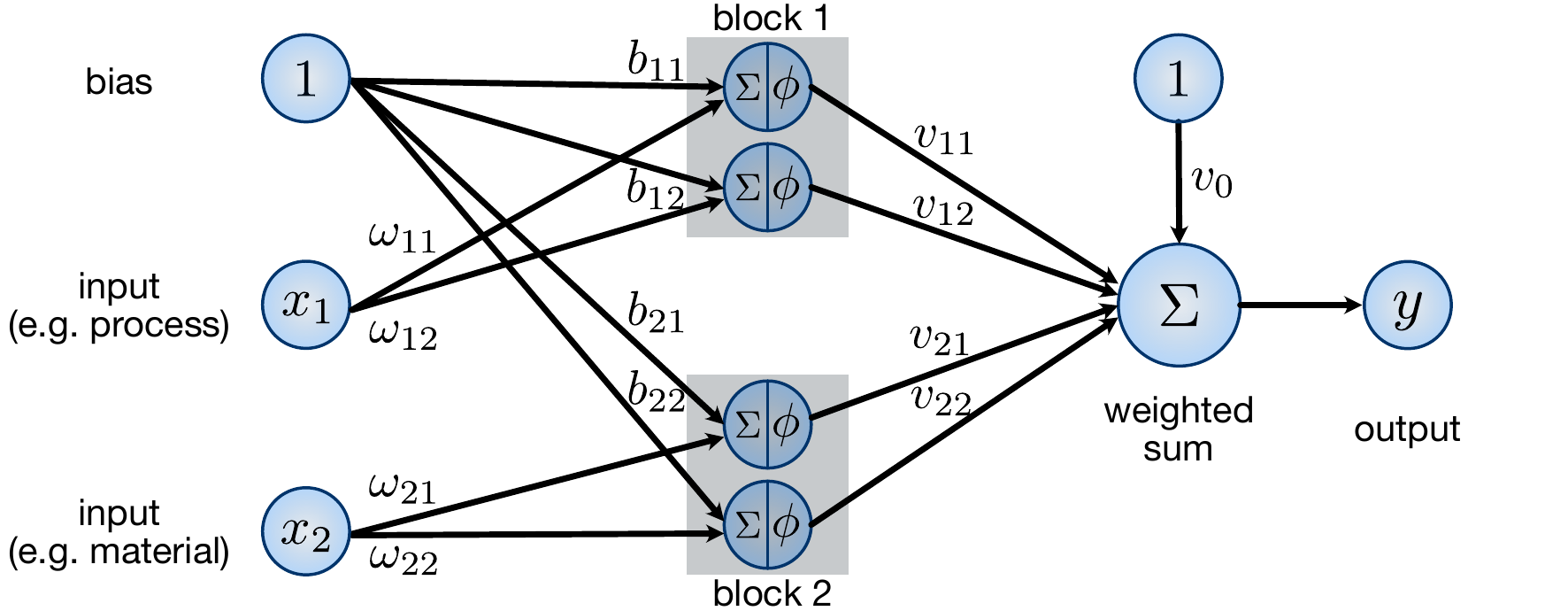}
\caption{Graphical representation of a blocked neural network with two neurons per block. In this sketch only two inputs are visualized, while the blocked neural network in this study has five inputs (two process and three material parameters).}
\label{fig:blocked_neural_network}
\end{figure}

A blocked neural network has one (so-called hidden) layer with blocks of neurons. All neurons in each block have -- besides the bias -- only one input parameter. We denote the input parameters by $x_i$, $i=1,\ldots,N$. The number of neurons in each block does not need to be equal. We denote the number of neurons in the block $i$ by $M_{i}$. The weight coming from the input $x_i$ to the $j$-th neuron in block $i$ is denoted by $\omega_{ij}$. Similarly, the weight of the bias to the $j$-th neuron in the $i$-th block is abbreviated by $b_{ij}$. The weights for the summation of the neurons in the hidden layer are denoted by $v_{ij}$, $i=1,\ldots,N$ and $j=1,\ldots,M_i$, respectively. With $v_0$ we denote the weight for the bias neuron. Then the output of this blocked neural network can be computed for the activation function $\phi$ as

\begin{linenomath}\begin{equation}
y = f_{\text{BNN}}\left(\mathbf{x}, \Theta\right) = v_0 + 
\sum_{i=1}^{N} \sum_{j=1}^{M_i} v_{ij} \phi\left( b_{ij} + \omega_{ij} x_i\right). \label{eq:bnn}
\end{equation}\end{linenomath}

Hereby, $\Theta$ summarizes all weights, i.e., the parameters of the blocked neural network, while $\mathbf{x}\in\mathbb{R}^{N}$ abbreviates the $N$ input parameters and $y\in\mathbb{R}$ the output parameter. The neuron activation function is chosen to be of sigmoidal type, i.e., $\phi(x)=\frac{e^x-e^{-x}}{e^x+e^{-x}}$. 

The weights have to be determined by the given input and output data. This is done by minimizing the mean squared error over a part of the given data, the so-called training set. The remaining data sets are used to validate the regression model. More precisely, the performance of the neural network is measured by the prediction mean squared error, which is estimated by cross validation (see \cite{efron:1993} for more details). 

Additionally, we like to analyze the sensitivity of the cause-and-effect-relations. Hereto, we compute the first partial derivative of the regression function with respect to each input parameter. A large value of the partial derivative indicates a large influence of the corresponding input parameter, i.e., small changes in the input will lead to large changes in the output. Furthermore, the sign of the partial derivative is important. A positive partial derivative indicates that an increase in the input leads to an increase in the output, while a negative sign means that an increase in the input leads to a decrease in the output. Now the advantage of the blocked neural network approach becomes clear. Computing the partial derivatives with respect to the input parameters can easily be done in such a neural network. Due to \cite{sarishvili:2006} we have: 
\begin{linenomath} \begin{equation} \label{eq:effect}
\frac{\partial f_{\text{BNN}}(\mathbf{x},\Theta)}{\partial x_i}=\sum_{i=1}^{M_i}v_{ii} \left(1-\phi\left(b_{ii} + \omega_{ii} x_i \right)^2\right) \omega_{ii}, \quad i=1,...,N
\end{equation} \end{linenomath}
where $M_i$ is the number of neurons in the $i$-th block as defined as in equation (\ref{eq:bnn}). 

Comparing the partial derivatives of different input parameters with each other is not so easy. A scalar quantity summarizing the cause-and-effect-relations would be desirable. A popular measure to quantify the sensitivity is the so-called average elasticity (AE). The average elasticity quantifies the percentage change of the output parameter with respect to a one percent change of the input parameters and thus is a dimensionless quantity. In practice, the average elasticity is computed for given samples of input and output data as follows. 

Let us assume that we have a total number of $S$ samples $(\mathbf{x}^{(s)}; y^{(s)})$ of input/output data and a blocked neural network $f_{\text{BNN}}$ as a nonlinear regression model approximating this data. Then the average elasticity for the $i$-th input parameter $x_i$ is defined (see \cite{refe:1996}) as
\begin{linenomath} \begin{equation} 
\text{AE}(x_i) = \frac{1}{S} \sum_{s=1}^{S} 
\left(\left|\frac{\partial f_{\text{BNN}}(\mathbf{x},\Theta)}{\partial x_{i}^{(s)}} \right| \right) 
\left( \left| \frac{x_{i}^{(s)}}{y^{(s)}} \right| \right), \quad
y^{(s)} \neq 0 \text{ for all } s = 1, \ldots, S.
\label{AE}
\end{equation} \end{linenomath}

\begin{figure}[!htb] \centering
\includegraphics[width=0.3\textwidth]{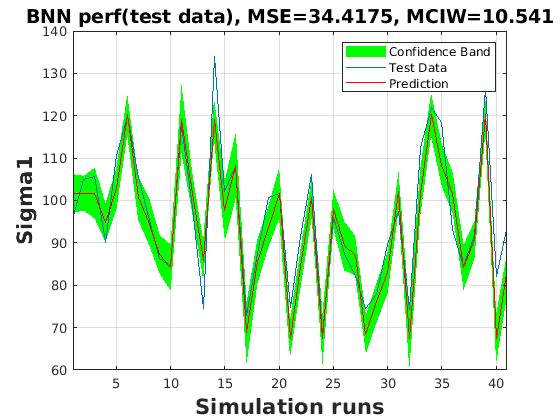}
\includegraphics[width=0.3\textwidth]{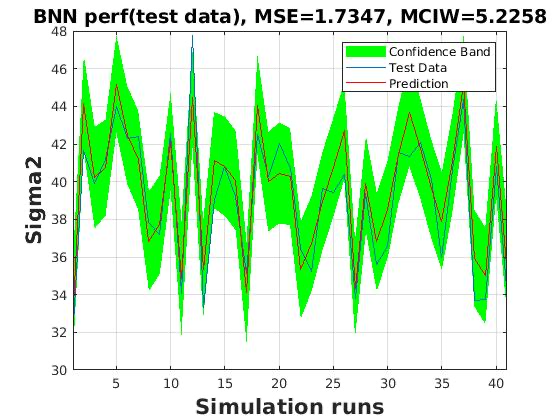}
\includegraphics[width=0.3\textwidth]{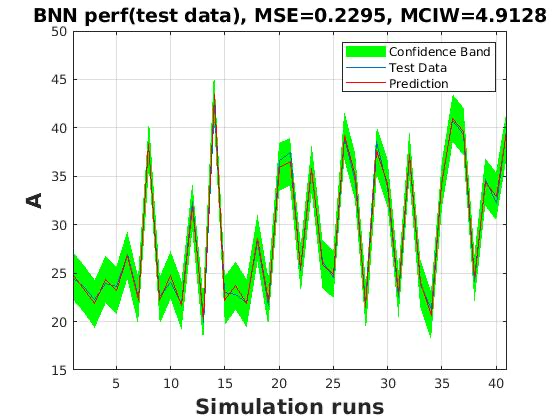}
\caption{Performance of the trained BNN for the three output parameters $\sigma_1$ (left), $\sigma_2$ (center), and $A$ (right).} 
\label{fig:performance}
\end{figure}

For our analysis, we choose a blocked neural network with two neurons in each of the blocks as presented in Fig.~\ref{fig:blocked_neural_network}. The weights of each block are trained on the training data set by the Levenberg-Marquardt Method \cite{hagan:1994}. We split the data set randomly in training and test data sets with a \SI{80}{\percent}-\SI{20}{\percent}-distribution, respectively.
For completeness, the performance of the training of the blocked neural network is visualized in Fig.~\ref{fig:performance}. Note that the confidence intervals have been estimated by the delta method described in \cite{papa:2000}.

\section{Conclusion}
With the trained blocked neural network we are now able to analyze the influence of the input parameters of the spunbond system on the output variables. Hereby, we distinguish the two categories of input parameters, i.e., the process parameters (spinning speed and pressure of the suction) and the material parameters (E modulus, density, and line density). The results are presented in the following figures for each output parameter $\sigma_1$, $\sigma_2$, and $A$. 

\begin{figure}[!htb] \centering
\includegraphics[width=0.49\textwidth]{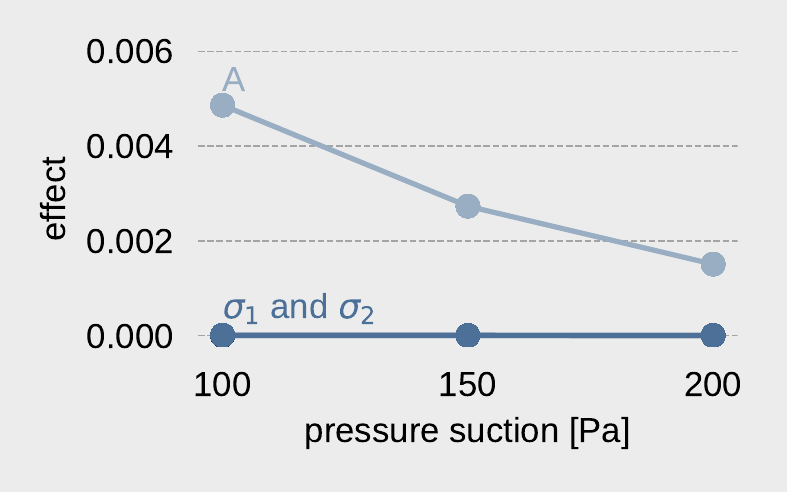} 
\includegraphics[width=0.49\textwidth]{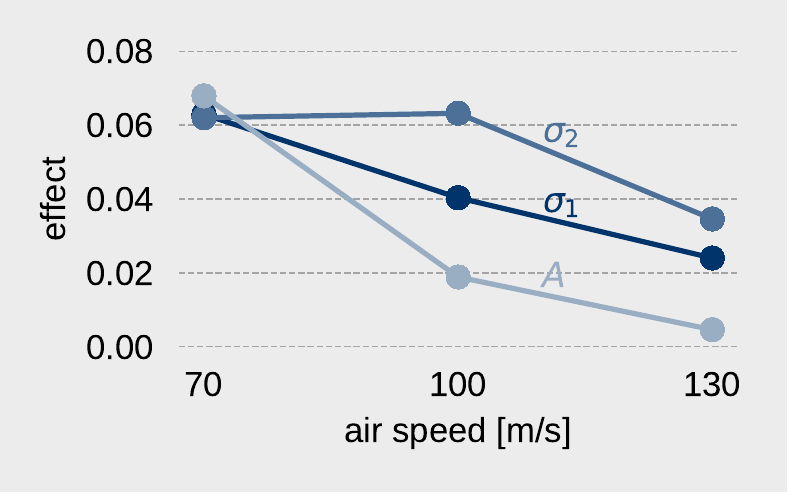}
\caption{Effect of the input process parameters on the fiber laydown characteristics: the pressure effect (left figure) is with \num{0} for $\sigma_1$/$\sigma_2$ and less than \num{0.005} for $A$ negligible. The effect on the air speed onto the fiber laydown characteristics (right figure) is with a range of \num{0.01} to \num{0.07} ten times bigger than the influence of the pressure.}
\label{fig:effect_process}
\end{figure}

\begin{figure}[!htb] \centering
\includegraphics[width=0.49\textwidth]{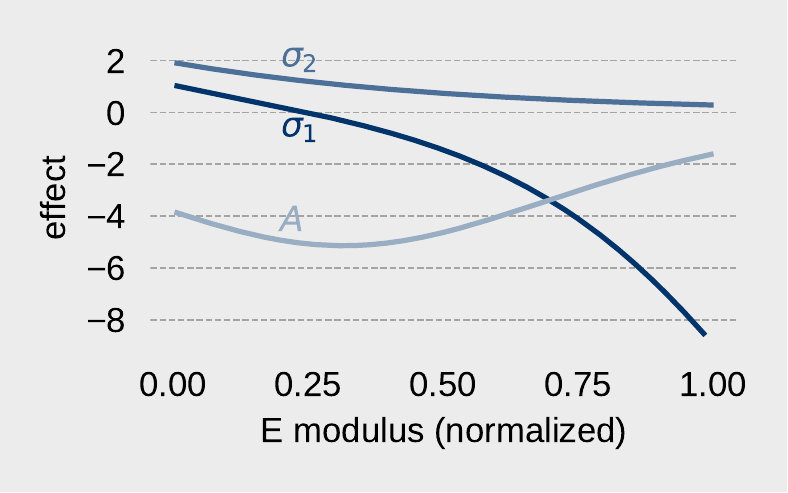} 
\includegraphics[width=0.49\textwidth]{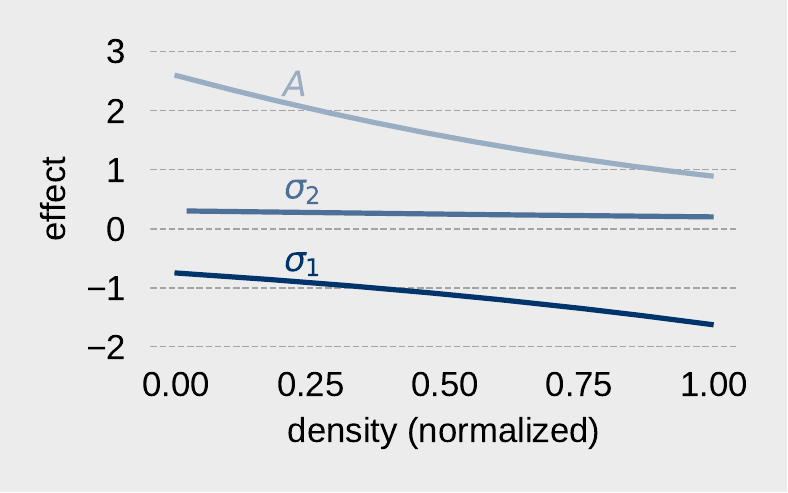} \\[0.05cm]
\includegraphics[width=0.49\textwidth]{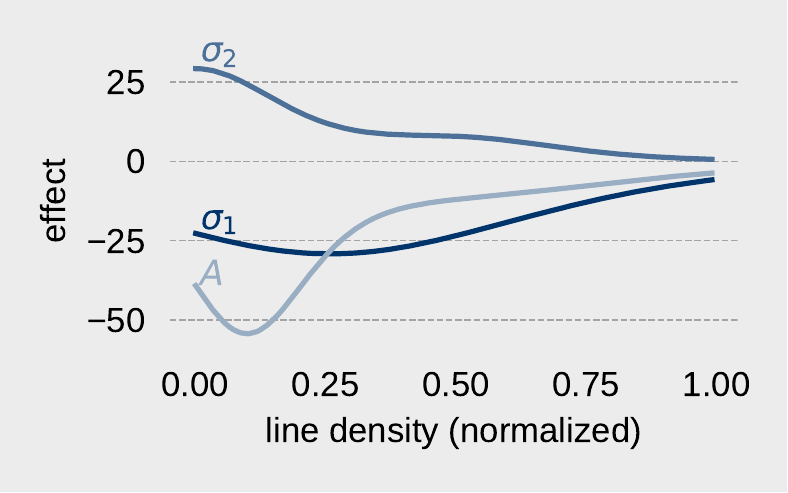}
\caption{Effect of the input material parameters on the fiber laydown characteristics: the effect of E modulus (first row, left figure) and the effect the density (first row, right figure) are on the same order of magnitude, while the effect of the line density is approximately 10 times bigger than the effect of E modulus and density (see second row, center figure).}
\label{fig:effect_material}
\end{figure}

We start our analysis with a comparison of the cause-and-effect relations computed due to equation (\ref{eq:effect}). First, we present the results for the input process parameters (see Fig.~\ref{fig:effect_process}). Then we visualize the effect results for the material parameters (see Fig.~\ref{fig:effect_material}). We conclude, that the effect of the process parameters is small. The material parameters clearly have more influence, in particular the line density has factor 10 more effect on the fiber laydown characteristics than E modulus and density. Furthermore, we observe the material input parameters have opposite effects in machine and cross machine direction. For example, an increasing density leads to an increasing $\sigma_2$, i.e. an increasing standard deviation of the throwing range in cross machine direction of the fiber laydown mass, while at the same time the standard deviation in machine direction $\sigma_1$ decreases. This behavior is also true for the cause-and-effect-relation of the line density. An increase of the line density leads to an increase of $\sigma_2$, but a decrease of $\sigma_1$. Surprisingly, the sign of the cause-and-effect-relation for E modulus vs. $\sigma_1$ changes. In lower regimes of the fiber's E modulus we observe the same effects, i.e., an increase of the E modulus leads to increased values of $\sigma_1$ and $\sigma_2$. But for higher values of the E modulus the throwing range in machine direction switches to decreased values of $\sigma_2$ for increased values of $E$.

Comparing the cause-and-effect relations with each other is not so easy. Hence, we introduced in section \ref{sec:04} a scalar measure -- the so-called average elasticity measure -- in order to quickly rank the effect of the input parameters on the fiber laydown. We compute the average elasticity measures due to equation (\ref{AE}). Table~\ref{AET} shows the average elasticity values for each of the five input parameters with respect to the three different output parameters. 

\begin{table}[htb!] \centering
\caption{Average elasticity measures (dimensionless quantities) computed due to equation (\ref{AE})}
\begin{tabular}{llll} \toprule
 & $\sigma_1$ & $\sigma_2$ & $A$ \\ \midrule
air speed         & \num{0.044}     & \num{0.134}     & \num{0.087}\\
pressure suction  & \num{0.0}       & \num{0.00001}   & \num{0.019}\\
E modulus         & \num{0.029}     & \num{0.007}     & \num{0.056}\\
density           & \num{0.007}     & \num{0.002}     & \num{0.022}\\ 
line density      & \num{0.136}     & \num{0.061}     & \num{0.203}\\ \bottomrule
\end{tabular}
\label{AET}
\end{table}

From the average elasticity measures in Tab.~\ref{AET} we conclude that the line density (titer) has the largest influence on the fiber laydown. Hereby, the impact on the fiber laydown in machine direction ($\sigma_1$) is more than doubled compared to the influence in cross machine direction ($\sigma_2$). As expected, changes of the air speed influence the fiber laydown as well. In this case, the impact on the cross machine direction is approximately three times bigger than the impact in machine direction. At first glance, the E modulus seems to be of minor impact, since the average elasticity measure is between \num{0.007} and \num{0.056} (relative change due to change of input), but a look at the effect plots shows an interesting effect that can not be detected by the summarized scalar quantity. As explained above, there is a change of the sign of the partial derivative in the cause-and-effect-relation, so in this case the summarized quantity of the average elasticity is not applicable. Again, the pressure of the suction is negligible.

\section{Summary and Outlook}
In this paper we present a mathematical and physical framework to simulate spunbond processes. Furthermore, we present a fiber laydown criterion that characterizes the laydown on the conveyor belt. We set up a design of experiments DoE for two process parameters of an academic spunbond process and three material parameters of the filaments. The simulation results in this DoE are analyzed by a blocked neural network. The BNN is not only used to predict the fiber laydown characteristics for the five input parameters, but also to analyze the cause-and-effect-relations with relevance plots. Additionally, the average elasticity measure leads to a quick ranking of the influencing effects with respect to their statistical significance.

The proposed simulation framework can be applied to optimize spunbond processes with respect to homogeneity of the fiber mass distribution on a conveyor belt. However, the considered framework addresses the laydown of one single fiber and its characterization in machine and cross machine direction. For future work the superposition of multiple fibers forming a 3D microstructure and its height distribution should be investigated. Therefore, a three-dimensional fiber laydown model based on stochastic differential equations can be used as described in \cite{klar:2012} or a construction of 3D nonwovens using a greedy approximation of the distribution of fiber directions as derived in \cite{gramsch:2018}.

\section*{Acknowledgements}
This work was developed in the Fraunhofer Cluster of Excellence “Cognitive Internet Technologies”. The authors would like to thank Walter Arne for performing the CFD simulations. The authors thank the anonymous referees for their valuable suggestions, which helped to improve the manuscript.

\section*{Conflict of Interests}
The authors declare that there is no conflict of interest regarding the publication of this paper.

\printendnotes

\bibliography{references}

\end{document}